\documentclass[10pt,conference,a4paper]{IEEEtran}
\pdfoutput=1
\usepackage[pdftex]{graphicx}
\usepackage{amsmath,amssymb}
\usepackage{tikz}
\usepackage{pgfplots}
\usepackage{xspace}
\usepackage{booktabs}
\pgfplotsset{compat=1.13}
\usepackage{url}
\usepackage{hyperref}

\usepackage{float}

\usepackage{amsmath,amsfonts}

\usepackage{algorithmic}
\usepackage{graphicx}
\usepackage{textcomp}
\usepackage{multirow}
\usepackage{xcolor}

\makeatletter
\DeclareRobustCommand\onedot{\futurelet\@let@token\@onedot}
\def\@onedot{\ifx\@let@token.\else.\null\fi\xspace}

\makeatother
\IEEEoverridecommandlockouts
\begin{document}
\title{SGDE: Secure Generative Data Exchange for Cross-Silo Federated Learning}

\author{\IEEEauthorblockN{Eugenio Lomurno\IEEEauthorrefmark{1}, Alberto Archetti\IEEEauthorrefmark{2}, Lorenzo Cazzella\IEEEauthorrefmark{3}, \\
Stefano Samele\IEEEauthorrefmark{4}, Leonardo Di Perna\IEEEauthorrefmark{5}, Matteo Matteucci\IEEEauthorrefmark{6}}
\vspace{5px}
\IEEEauthorblockA{Politecnico di Milano, Italy
\vspace{5px}
\\
Email: eugenio.lomurno@polimi.it\IEEEauthorrefmark{1},
alberto.archetti@polito.it\IEEEauthorrefmark{2},
 lorenzo.cazzella@polimi.it\IEEEauthorrefmark{3},\\ stefano.samele@polimi.it\IEEEauthorrefmark{4},
 leonardo.diperna@polimi.it\IEEEauthorrefmark{5},
 matteo.matteucci@polimi.it\IEEEauthorrefmark{6}}
}

\maketitle

\begin{abstract}
Privacy regulation laws, such as GDPR, impose transparency and security as design pillars for data processing algorithms. In this context, federated learning is one of the most influential frameworks for privacy-preserving distributed machine learning, achieving astounding results in many natural language processing and computer vision tasks. Several federated learning frameworks employ differential privacy to prevent private data leakage to unauthorized parties and malicious attackers. Many studies, however, highlight the vulnerabilities of standard federated learning to poisoning and inference, thus raising concerns about potential risks for sensitive data. To address this issue, we present SGDE, a generative data exchange protocol that improves user security and machine learning performance in a cross-silo federation. The core of SGDE is to share data generators with strong differential privacy guarantees trained on private data instead of communicating explicit gradient information. These generators synthesize an arbitrarily large amount of data that retain the distinctive features of private samples but differ substantially. In this work, SGDE is tested in a cross-silo federated network on images and tabular datasets, exploiting beta-variational autoencoders as data generators. From the results, the inclusion of SGDE turns out to improve task accuracy and fairness, as well as resilience to the most influential attacks on federated learning.
\end{abstract}

\IEEEpeerreviewmaketitle

\section{Introduction}
\label{sec:introduction}

The formal definition of strict privacy regulation laws, such as the European GDPR~\cite{albrecht_how_2016} and the Chinese Cyber Security Law~\cite{parasol_impact_2018}, raised the need to impose the fundamental right of privacy as a design pillar for data elaboration algorithms. Today, it is of utmost importance for AI researchers and developers to provide sound and secure algorithms that minimize the risks for data owners while providing value and knowledge. 

Federated Learning (FL)~\cite{li_federated_2020} is among the most popular frameworks for distributed machine learning, born with a strong commitment to privacy preservation. In FL, a set of clients, each holding private data samples, collaborate to train a single machine learning model with the help of a central server. In the original algorithm, FedAvg, developed by McMahan \textit{et al.}~\cite{mcmahan_communication-efficient_2017}, the central server initializes and broadcasts a shared model to a set of clients. Then, each of these clients performs a small number of Stochastic Gradient Descent (SGD) epochs on their private data. The updated weights of the model are then sent back to the central server and averaged proportionally to the number of samples involved in the local optimization steps. Finally, the central server broadcasts the aggregated model and repeats the process until convergence. 

During the training procedure, private data samples never leave the clients; therefore, in principle, user privacy is preserved. 
However, the iterative exchange of model weights exposes an attack surface that can be exploited by members inside the federation with malicious intents.
Indeed, a set of poisoning and inference techniques based on generative deep learning methods have been developed to retrieve secret information.
For instance, many inference attacks manipulate the gradients sent to the server at each iteration~\cite{lyu_threats_2020} to reconstruct private data or alter the central model with carefully designed updates.

The growing popularity of attacks on the secrecy of private data has questioned the practical security level of federated learning.
Thus, defense against attacks that threaten privacy is one of the biggest open problems for FL research.
To this end, many defensive techniques have been developed to mitigate threats, e.g., robust model aggregation~\cite{pillutla2019robust}, model pruning~\cite{caldas2018expanding}, and gradient encryption~\cite{zhang2020batchcrypt}. 
Unfortunately, a complex and large-scale system, like the FL framework, is exposed to unique vulnerabilities and risks, and often, the adopted countermeasures are not adequate to ensure robust security standards~\cite{9411833}.

\begin{figure}[t]
    \centering
    \includegraphics[width=\columnwidth]{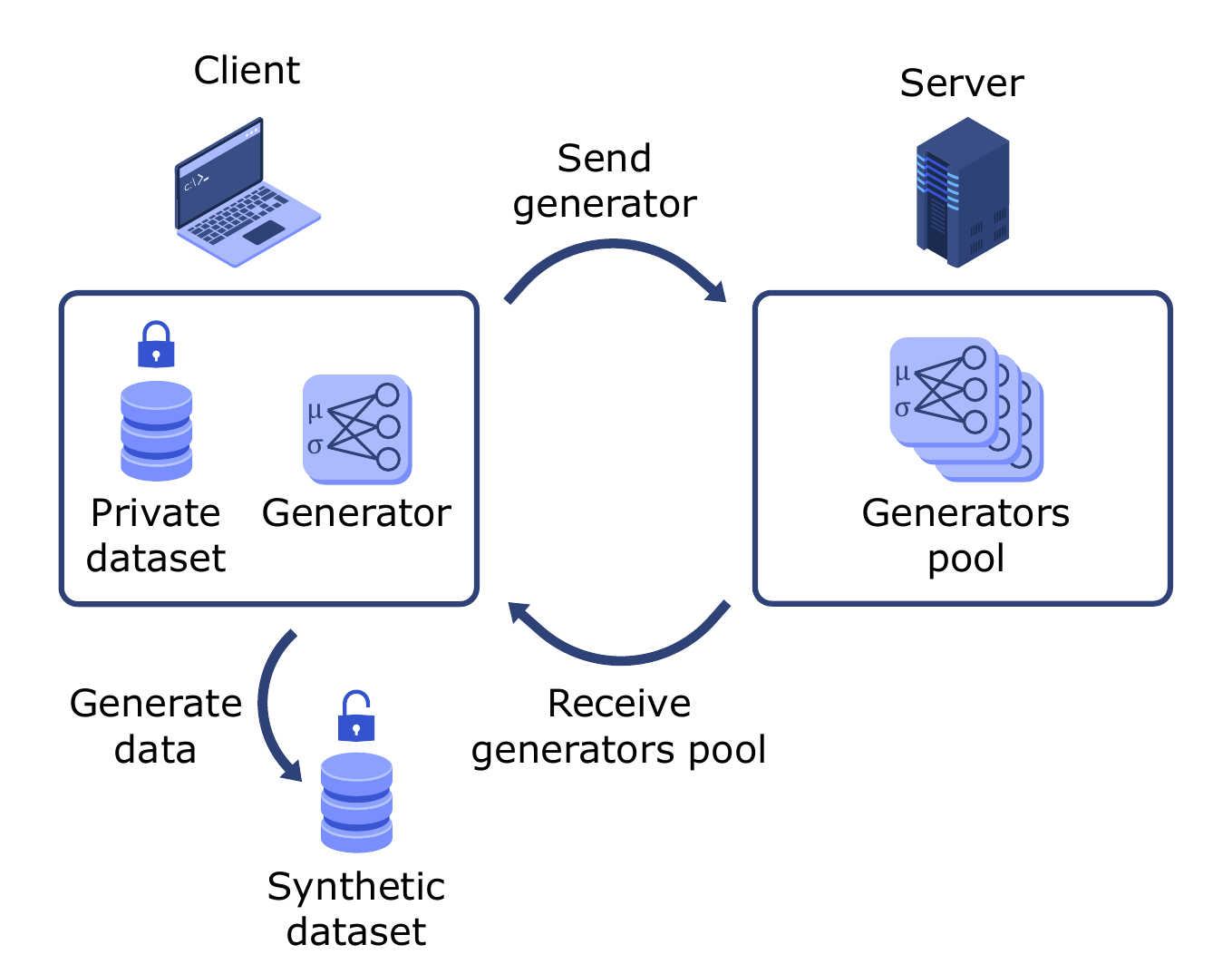}
    \caption{The general scheme of SGDE from a single client perspective. In exchange for a generator trained on private data, the server grants the client access to the entire generators pool. Then, the client can generate an arbitrarily large synthetic dataset for offline use.}
    \label{fig:sgde_intro}
\end{figure}

To improve safety and grant privacy of user data in a federated context, we propose SGDE, a framework for secure data exchange through differentially private data generators (Figure~\ref{fig:sgde_intro}). 
SGDE operates in three phases: \emph{Subscribe}, \emph{Push}, and \emph{Pull}. In the \emph{Subscribe} phase, the client communicates to the server its intention to take part in the data exchange process. 
In the \emph{Push} phase, the client trains a set of data generators with a high differential privacy level following the constraints prescribed by the server and sends the data generators to the server.
Finally, in the \emph{Pull} phase, the client may access the set of generators stored by the server and train any machine learning model on generated data locally. 
This study is focused on the cross-silo federated learning setting, where clients are institutions, e.g., hospitals, universities, or companies.
Cross-silo federated learning is characterized by a low number of clients -- hundreds, at most -- and each client has access to the computational power needed to train a generative model locally.

SGDE enables each client to generate an arbitrarily large set of synthetic data that preserves the distinctive features of real data.
Synthetic data is available offline for client-side machine learning, combining the flexibility and transparency of a centralized dataset while reducing communication costs.
Also, in a supervised classification setting, SGDE enables equal generation of samples for each class, increasing fairness with respect to underrepresented classes in the real dataset.

We argue that SGDE is secure against the most prominent attacks to federated learning, namely, poisoning and inference.
In fact, from the security standpoint, an offline synthetic dataset is more advantageous than a privacy-protected distributed dataset, as poisoning attacks are much easier to detect~\cite{fung_mitigating_2020}.
Also, since there is no single model to be poisoned nor a global task to learn, the attack surface for malicious agents is strongly reduced, mitigating attacks by design, and enhancing the framework security.

This work brings the following contributions: 
\begin{itemize}
    \item we propose the SGDE framework and extensively discuss the advantages of its inclusion in a federated setting; 
    \item we discuss how its implementation through differential-privacy-compliant data generators translates into remarkable improvements from a security standpoint; 
    \item our claims about SGDE are validated by experimentally testing the framework with tabular and images datasets, demonstrating performance improvements from the federation members point of view; 
    \item the SGDE framework demonstrates capabilities to deal with distribution biases and discriminations. 
\end{itemize}

The rest of this work is organized as follows.
Section~\ref{sec:related_work} collects the most recent works related to federated networks, differential privacy, and generative models, highlighting  the motivations behind the development of SGDE. 
Section~\ref{sec:method} presents SGDE from a formal point of view, providing a detailed description of the protocol and analyze its security advantages.
In Section~\ref{sec:experiments} collects the experiments to validate the claims about SGDE and extensively discuss the results.
Finally, Section~\ref{sec:conclusion} concludes this work by summarizing the main concepts presented.

\section{Related Work}
\label{sec:related_work}

This section collects the most influential works related to federated learning and privacy preservation in the machine learning context.
In particular, Section~\ref{sec:federated_learning} presents Federated Learning (FL) as the target scenario for the application of SGDE.
Then, Section~\ref{sec:differential_privacy} describes Differential Privacy (DP) and its relation to FL security alongside an overview of the main techniques in generative deep learning and how they relate to DP.
Finally, Section~\ref{sec:threats_to_fl} describes the most prominent attacks to the FL paradigm and the secrecy of user data.

\subsection{Federated Learning}
\label{sec:federated_learning}

Federated Learning (FL)~\cite{li_federated_2020,kairouz_advances_2021} is a privacy-compliant framework for distributed machine learning, scalable to millions of devices.
In the general setting, $K$ clients, each equipped with a local dataset of private samples, collaborate with a central server to minimize a global loss function $F$ on a model with parameters $w$:
\begin{equation}
    \min_w F(w)=\min_w\sum\limits_{k=1}^K p_k F_k(w).
    \label{eq:fl}
\end{equation}
In~\eqref{eq:fl}, $F_k$ is the loss function of client $k$ evaluated on $n_k$ samples from the local dataset. 
$p_k$ is a weighting factor, such that $p_k \geq 0$ and $\sum\nolimits_{k=1}^K p_k = 1$. 
The values of $p_k$ depend on the application, but the most common assignments, given $n=\sum\nolimits_{k=1}^K n_k$, are $p_k = \frac{n_k}{n}$ and $p_k = \frac{1}{K}$.

The main challenge of FL is dealing with the many faces of heterogeneity to be found in a massive network of clients. 
First, computational capabilities and connectivity may vary a lot between different clients. 
Indeed, the federated network may comprise devices with different hardware architectures and memory constraints.
Also, the communication channel may be unreliable, leading to long pending periods and lost updates.
In this setting, naively ignoring dropped updates may introduce a bias in the trained model towards the features of clients with more reliable connectivity.
To this end, many efforts have been devoted to reducing the communication cost in FL systems and to dealing with dropped updates~\cite{konecny_federated_2017,sattler_robust_2020}. 
As stated in~\cite{lim_federated_2020}, the main techniques to reduce the required bandwidth in mobile networks are increased edge computation, model compression, and importance-based updating.
Energy efficiency is another topic of interest, as wireless communication is extremely power-hungry, especially in network federations including IoT and battery-powered devices~\cite{yang_energy_2021,guler_sustainable_2021}.

Another dimension of heterogeneity is the statistical diversity between the data distribution of each client, making the IID assumption unrealistic in real-world FL scenarios. 
Many techniques have been introduced to deal with federated data heterogeneity.
Agnostic FL~\cite{mohri_agnostic_2019} optimizes the central model with respect to any mixture of client updates, naturally increasing the fairness of the trained model.
FedProx~\cite{li_federated_heterogeneous_2020} generalizes the original FedAvg algorithm~\cite{mcmahan_communication-efficient_2017} by adding a regularization term that guarantees convergence in the non-IID setting. 

The common FL framework provides a single model for each user, limiting the performance on local inference.
Many researchers advocate for personalization of local models as a method to deal with non-IID data distributions in federated networks~\cite{deng_adaptive_2020}.
In~\cite{wu_personalized_2020}, the authors address heterogeneity in a network of IoT devices and provide the main techniques to implement personalized on-device learning, namely, transfer learning, meta learning, federated multi-task learning, and federated distillation.
Concerning meta learning, most works~\cite{fallah_personalized_2020,jiang_improving_2019} rely on the MAML framework~\cite{finn_model-agnostic_2017} to provide a personalized model for each user.
In~\cite{huang_personalized_2021} attentive message passing facilitates personalization by aggregating clients with similar features.

\subsection{Differential Privacy and Generative Models}
\label{sec:differential_privacy}

Differential Privacy (DP)~\cite{dwork_algorithmic_2013} is a mathematically rigorous procedure to measure the security of a system against the disclosure of sensitive information related to individual samples.
In particular, a randomized mechanism $\mathcal{M}:\mathcal{X} \rightarrow \mathcal{Y}$ is $(\varepsilon)$-differentially private if for any pair of adjacent inputs $x, y \in \mathcal{X}$ and any subset of outputs $S \subseteq \mathcal{Y}$ the following holds:
\begin{equation}
    \text{Pr}[\mathcal{M}(x) \in S] \le \exp^\varepsilon \text{Pr}[\mathcal{M}(y) \in S].
    \label{eq:dp1}
\end{equation}
In~\eqref{eq:dp1}, $\varepsilon$ represents the privacy budget, i.e., the theoretical amount of information that could leak from the system.
The lower the $\varepsilon$ value, the stronger the privacy guarantee is. 
Moreover, DP exhibits three convenient properties that make its inclusion natural in iterative optimization procedures such as stochastic gradient descent:
\begin{itemize}
    \item \textbf{Composability}: a system composed by several differentially private mechanisms is still differentially private; this property holds for sequential and parallel composition.
    \item \textbf{Group privacy}: privacy guarantees never degrade abruptly, even if the adjacent inputs are strongly correlated or belong to the same individual.
    \item \textbf{Robustness to auxiliary information}: the privacy level is theoretically granted regardless of the knowledge the attacker has about the mechanism.
\end{itemize}

However, the constraints of $\varepsilon$-DP are too strict to make it viable for real world applications. To answer this issue, many relaxed variants have been developed, such as $\textit{f}$-DP~\cite{dong2019gaussian}, concentrated DP~\cite{dwork2016concentrated} and R\'{e}nyi DP~\cite{mironov2017renyi}. The most common relaxation of $\varepsilon$-DP is ($\varepsilon,\delta$)-DP~\cite{abadi_deep_2016}. In $(\varepsilon,\delta)$-DP, the randomized mechanism $\mathcal{M}:\mathcal{X} \rightarrow \mathcal{Y}$ is $(\varepsilon,\delta)$-differentially private if for any pair of adjacent inputs $x, y \in \mathcal{X}$ and any subset of outputs $S \subseteq \mathcal{Y}$ the following holds:
\begin{equation}
    \text{Pr}[\mathcal{M}(x) \in S] \le \exp^\varepsilon \text{Pr}[\mathcal{M}(y) \in S] + \delta.
    \label{eq:dp2}
\end{equation}
In~\eqref{eq:dp2}, the additive term $\delta$ represents the (possibly small) probability that $\varepsilon$-DP could be violated.

Many Federated learning systems include $(\varepsilon,\delta)$-DP-based techniques in the distributed learning process to increase the privacy level of private data stored in the client devices. 
A common technique is to define a $(\varepsilon,\delta)$-differentially-private optimizer, such as differentially-private SGD~\cite{abadi_deep_2016}, where DP standards are met by clipping and adding Gaussian noise, at each iteration, to the current gradient.
PATE~\cite{papernot_semi-supervised_2017} is a DP-aware training framework based on the student-teacher model, where a central model, the student, learns from a set of black-box private models, the teachers, by predicting outputs chosen with noisy voting.

DP countermeasures to data disclosure are also adopted in synthetic data generation through deep generative models, such as Generative Adversarial Networks (GANs)~\cite{goodfellow_generative_2014} and Variational Autoencoders (VAEs)~\cite{kingma_auto-encoding_2014}.
Differentially-private VAEs~\cite{chen_differentially_2018} and GANs~\cite{xie_differentially_2018} achieve DP standards by adding carefully designed perturbations to the gradients during training. 
Advanced GAN architectures have been extended to meet DP standards, such as InfoGAN~\cite{mugunthan_dpd-infogan_2021} and DP-Conditional GAN~\cite{torkzadehmahani_dp-cgan_2020}, relying on the R\'{e}nyi-DP model~\cite{mironov2017renyi,mironov2019r}.

DP-auto-GAN~\cite{tantipongpipat_differentially_2020} is a framework for synthetic data generation, applicable to unlabeled, mixed-type data, that combines the flexibility of GANs with the dimensionality-reduction capabilities of autoencoders. 
PATE-GAN~\cite{yoon2018pategan} is an extension of the PATE framework in which a set of private teacher discriminators train a student discriminator against a common generator. 
In this way, synthetic data is differentially private with respect to the original, private data.
In~\cite{acs_differentially_2018}, the authors separate data synthesis in two steps.
First, they perform $K$-means clustering with a differentially private kernel over sensitive data.
Then, they train $K$ generative models, one for each cluster, achieving higher data utility than a single-model architecture.

\subsection{Threats to federated learning}
\label{sec:threats_to_fl}

A federated learning pipeline usually involves differential privacy techniques to protect clients data against unintended disclosure.
However, many authors show that targeted attacks can retrieve sensitive information, even when differential privacy is involved.
The most influential attacks to federated learning are collected in~\cite{lyu_threats_2020} and categorized as poisoning attacks and inference attacks.
During a poisoning attack, a malicious agent deliberately modifies the training data or the parameters of the local model to deviate the learning procedure away from the true objective.

Inference attacks, instead, aim at guessing whether a specific data sample took part in the training procedure (membership inference~\cite{shokri_membership_2017,hayes_logan_2018}) or reconstructing the input sample given a model and its corresponding output (input inference or model inversion~\cite{fredrikson_model_2015}).

Another attack surface regards the backbone of FL training, the iterative gradient exchange.
These inference methods are called gradient leakage attacks and allow malicious agents to obtain information about raw private data samples with GAN-based gradient reconstruction~\cite{melis2019exploiting}.
It has been shown that even small portions of intermediate updates can lead to sensitive local data disclosure~\cite{aono2017privacy}.

\section{SGDE: Secure Generative Data Exchange}
\label{sec:method}

This section introduces SGDE, a data exchange framework based on generative models.
The goal of SGDE is to guarantee strong privacy levels and face the major security issues related to the Federated Learning (FL) domain.
It is important to note that SGDE is not a \emph{learning} protocol, as there is no target model to be optimized nor a predefined task to be solved.
Instead of sharing models gradients, SGDE provides to each client in a cross-silo federation a set of differentially private data generators able to synthesize an arbitrarily large number of samples.
Each data generator embodies the features of its corresponding private dataset without disclosing any explicit information to curious or malicious agents.
Once the generators are shared, each client can freely generate synthetic samples for any local machine learning task.
In fact, the purpose of the dataset is entirely up to the client.
Machine learning can occur privately, directly on generated data, or the client may still participate in a federated iterative procedure, training a common model in a distributed fashion using synthetic data, with clear privacy advantages, since synthetic data does not retain any sensitive information.

This work is focused on supervised classification, but the SGDE framework can be easily extended to other machine learning tasks as well. We argue that SGDE brings the following advantages to a cross-silo federation:
\begin{itemize}
    \item \textbf{Flexibility}: SGDE gives full control of synthetic data to the client, both from the generation and usage aspects. The client can choose the best generators to build the dataset and arbitrarily decide the dataset cardinality. Moreover, synthetic data samples from SGDE are task-agnostic and can be involved in different machine learning applications, without the need for a central authority to coordinate the federation parties.
    \item \textbf{Security}: in SGDE, private data never leave the client, and the message exchange involves only generators. Also, private training parameters remain local, except for the generator privacy level. In this way, the attack surface with respect to poisoning and inference is severely limited.
    \item \textbf{Fairness}: each client can generate an arbitrarily large number of samples for each predefined label. In this way, each class can populate the dataset in equal proportion. This means that a synthetic dataset is fair with respect to class representation so that, as shown in Section~\ref{sec:experiments}, it is possible to attenuate distribution biases.
    \item \textbf{Communication efficiency and robustness}: SGDE does not involve any iterative exchange of data, as training occurs entirely on the device with no internet communication involved. Instead, generators are exchanged once between clients and the central server, providing a huge advantage in terms of bandwidth usage.
\end{itemize}

\begin{figure}[t]
    \centering
    \includegraphics[width=\columnwidth]{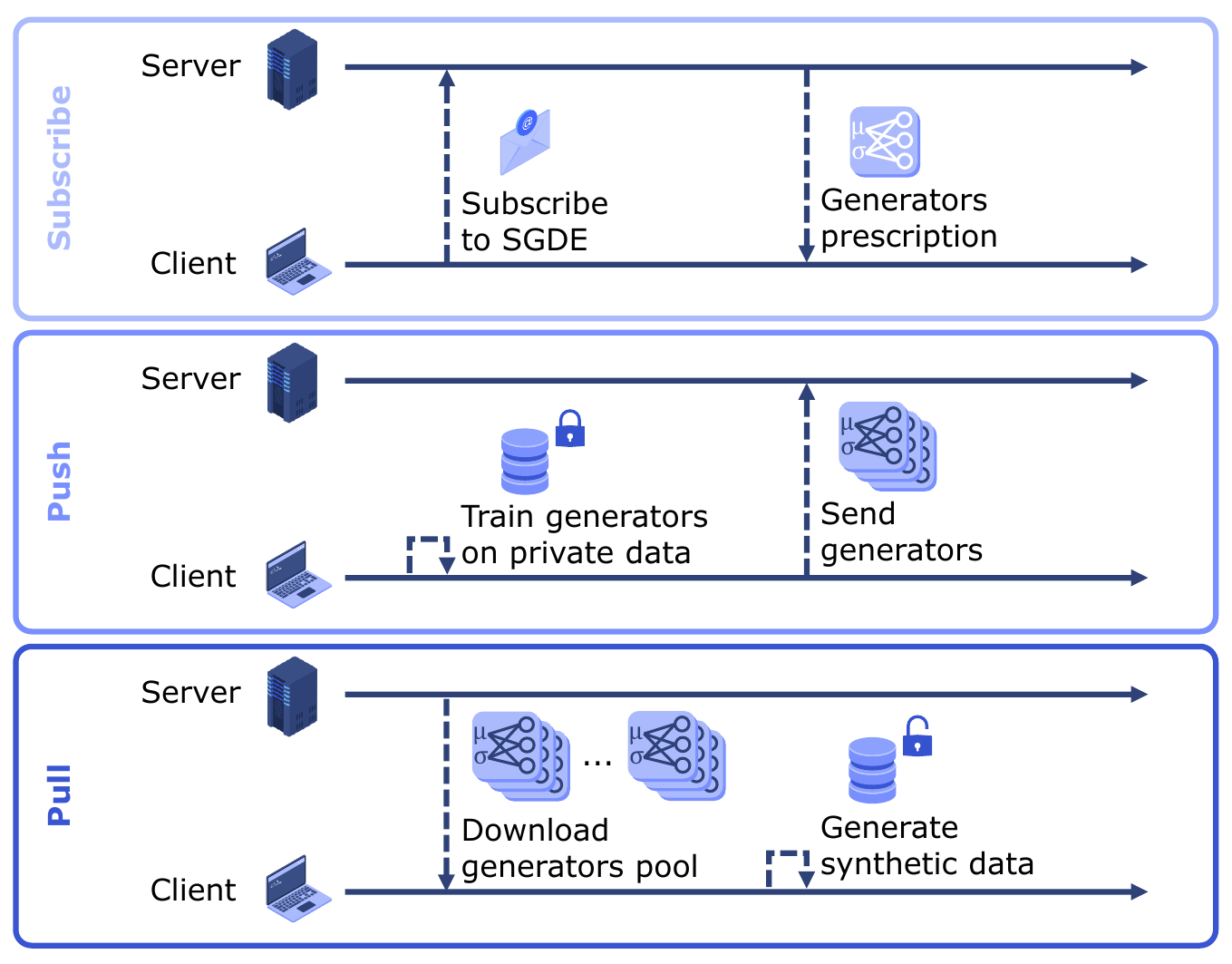}
    \caption{The three steps of SGDE. First, during the \emph{Subscribe} phase, client and server exchange preliminary information. Then, in the \emph{Push} phase, the client trains a generator on private data according to the server prescriptions and sends it to the server. Finally, in the \emph{Pull} phase, the client accesses the pool of generators available on the server. }
    \label{fig:sgde_prot}
\end{figure}

\subsection{The Steps of SGDE}
\label{sec:steps}

Consider a federated network of $K$ clients, such that each client $k$ is equipped with a private set of data samples such that $\mathcal{D}_k$ = $\{x_1, x_2, \dots, x_{n_k}|x_i \in \mathcal{X}\}$, where $\mathcal{X}$ corresponds to the data domain.
From now on, we assume to work in a supervised classification setting, but SGDE can be extended to other machine learning tasks. 
In this case, each sample $x_i$ is paired with a class label $y_i$ belonging to a finite set $\mathcal{Y}$.
Furthermore, we assume each client $k$ to be interested in accessing a larger set of data samples than $\mathcal{D}_k$, and willing to join a federation with other clients with the same intent.

In this context, SGDE is a highly secure solution for the federation described above 
In SGDE, each client, starting from its private dataset $\mathcal{D}_k$, is required to build a set of generators $\mathcal{G}_k$ composed of one data generator $g^y_k$ for each class $y\in\mathcal{Y}$.
Then, the set of generators $\mathcal{G}_k$ is collected by the central server and published among the other clients.

SGDE, as shown in Figure~\ref{fig:sgde_prot}, is articulated in three steps, namely, \emph{Subscribe}, \emph{Push}, and \emph{Pull}:

\begin{itemize}
    \item \emph{Subscribe}: client $k$ communicates to the central server its intention to join the protocol. The server responds with a set of requirements for the generators. These may include specifications regarding synthetic data, the internal structure of the generators, or the minimum level of $(\varepsilon,\delta)$-DP to guarantee.
    \item \emph{Push}: for each class $y \in \mathcal{Y}$, client $k$ trains a generator $g^y_k:\mathcal{Z}\rightarrow\mathcal{X}$, such that $g^y_k$ is able to synthesize samples $\hat{x}$ corresponding to label $y$, starting from a noise vector $z \in \mathcal{Z} \sim \mathcal{N}(\mathbf{0},\mathbf{\Sigma})$. Each $g^y_k$ is collected in a generator set $\mathcal{G}_k$ and sent to the central server, alongside the $(\varepsilon,\delta)$-DP level measured at the end of the training. Any other information related to the training procedure that occurred on client $k$ is not required by the server and should remain private.
    \item \emph{Pull}: client $k$ is granted access to the generators pool stored in the central server, containing generators from different clients. Client $k$ may access the parameters, structure, and privacy level of the generators.
    At this point, client $k$ may select which generators to download and start building a synthetic dataset. 
    Each generator can be used to sample an unbounded number of synthetic instances and to produce arbitrarily large datasets.
\end{itemize}

\subsection{Threat Analysis}
\label{sec:threats}

This section analyzes how SGDE resists to the most common attacks to federated learning.
Considering the topology of attacks described in~\cite{lyu_threats_2020}, the attention is focused is on attacks run by malicious agents inside the federation.  

The first family of attacks to be analyzed is poisoning. 
In SGDE, the attack surface for a malicious agent resides only in the construction of the generators. 
Since there is no central model nor a single point of failure, an attacker cannot compromise the whole system, as in a centralized FL setting.
At most, the attacker may build a set of generators that synthesize poisoned data with the intent of having local models produce wrong estimations.
However, from an honest user perspective, a poisoning attack can be detected by performance drops and solved by discarding malicious data generators during the synthetic data sampling process.
More generally, as argued in~\cite{fung_mitigating_2020}, a centralized and accessible dataset, such as a synthetic one generated with SGDE, is less prone to poisoning, as direct heuristic analysis makes poisoning attacks easier to detect. This is not possible in a standard FL scenario, where no data information is transparent to the clients, other than their private datasets.

The second family of attacks to be considered is inference.
Again, the attack surface of SGDE is much smaller with respect to a standard FL setting. 
The absence of iterative gradient exchange among the parties prevents attackers from carrying out gradient leakage attacks and reconstructing private data.

We argue that other inference threats over data generators, such as membership inference and model inversion attacks, would not be effective either with SGDE.
In fact, differential privacy was introduced in machine learning scenarios precisely as a countermeasure to these kinds of attacks.
As discussed in~\cite{rahman2018membership}, training models with strict DP levels guarantees high resilience against membership inference attacks.
At the same time, current model inversion attacks against differentially private data generators exchanged through SGDE turn out to be ineffective. 
Since the data generators map random noise to synthetic samples, a model inversion attack would reconstruct the latent noise, at most.

Outsider attacks are generally considered a threat to the network infrastructure rather than a threat to the learning protocol.
However, SGDE narrows the attack surface concerned with system availability too.
Reducing the communication to the exchange of generators decreases the amount of information to be sent over the internet, with an immediate benefit in terms of bandwidth usage.
This data flow reduction allows the SGDE framework to evade a broad set of server availability attacks concerning the continuous gradient exchange.

As a final note, we argue that traditional FL and SGDE can be symbiotically exploited to further improve security guarantees for individuals.
In fact, a potentially distributed learning system, training on a dataset generated via SGDE, should not be concerned about the secrecy of synthetic data.
Private users samples never leave the clients, while new machine learning algorithms can run on public synthetic data retaining the distinctive features of the private counterparts.


\section{Experiments}
\label{sec:experiments}

This section collects the set of experiments involving SGDE to validate our claims. 
In particular, we argue that the inclusion of SGDE in a machine learning scenario with private distributed data is beneficial from the utility, security, and fairness standpoints.
To this end, well-known machine learning models are evaluated according to the accuracy, F1 score, and AUC metrics in three different scenarios called \emph{Local}, \emph{Federated}, and \emph{Synthetic}.
In the \emph{Local} scenario, clients have no access to public generators and rely only on their private data to produce machine learning models locally.
In the \emph{Federated} scenario, a single machine learning model is jointly trained using the FedAvg algorithm~\cite{mcmahan_communication-efficient_2017} starting from private data.
Finally, in the \emph{Synthetic} scenario, clients take part in the SGDE protocol and exchange data generators to access a larger synthetic dataset.
In this case, local models are trained on synthetic data coming from generators provided by different clients.

Our experiments include five tabular datasets from the UCI Machine Learning repository~\cite{Dua:2019} (Titanic, Breast Cancer Wisconsin - Diagnostic, Mushroom, Adult, and Wine Quality) and two image datasets (MNIST~\cite{lecun2010mnist} and Fashion MNIST~\cite{DBLP:journals/corr/abs-1708-07747}).
Datasets with already defined training and test splits are kept unchanged, while for the other datasets a 90\%-train-10\%-test random split is applied.
The federation network is composed of 20 clients, each equipped with a 5\% non-overlapping random split of the original training dataset. 

\begin{figure*}[t]
  \centering
    \includegraphics[width=0.094\textwidth]{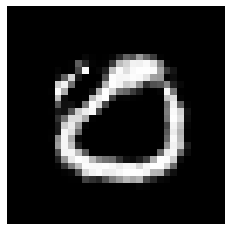}
    \includegraphics[width=0.094\textwidth]{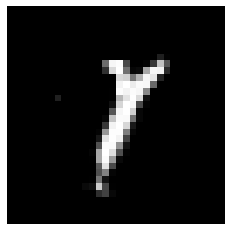}
    \includegraphics[width=0.094\textwidth]{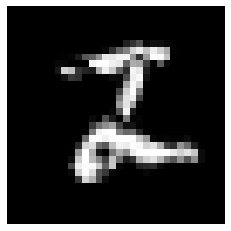}
    \includegraphics[width=0.094\textwidth]{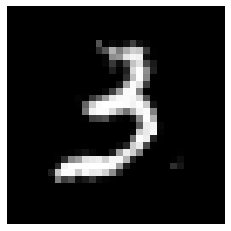}
    \includegraphics[width=0.094\textwidth]{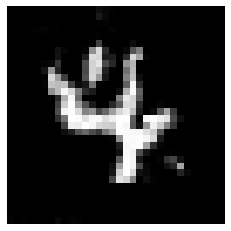}
    \includegraphics[width=0.094\textwidth]{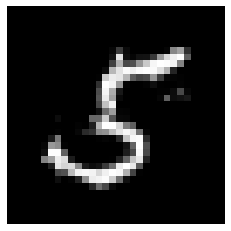}
    \includegraphics[width=0.094\textwidth]{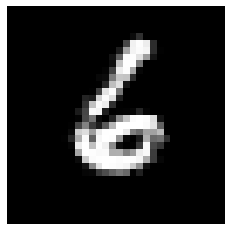}
    \includegraphics[width=0.094\textwidth]{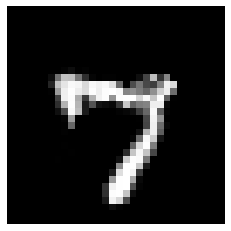}
    \includegraphics[width=0.094\textwidth]{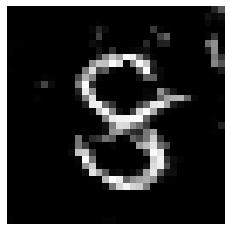}
    \includegraphics[width=0.094\textwidth]{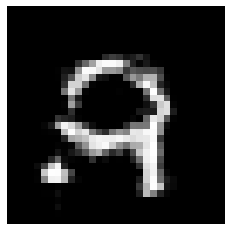}
    
    \includegraphics[width=0.094\textwidth]{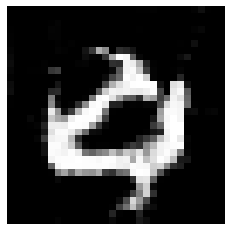}
    \includegraphics[width=0.094\textwidth]{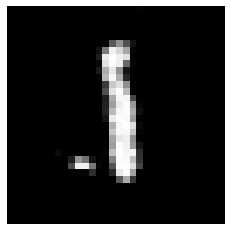}
    \includegraphics[width=0.094\textwidth]{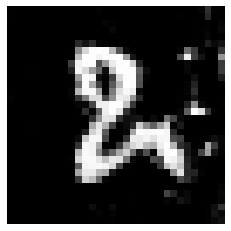}
    \includegraphics[width=0.094\textwidth]{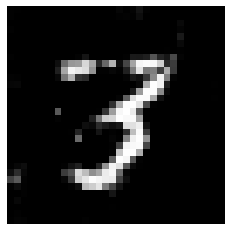}
    \includegraphics[width=0.094\textwidth]{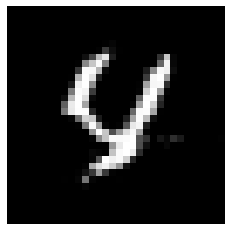}
    \includegraphics[width=0.094\textwidth]{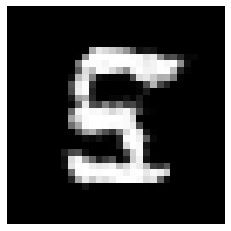}
    \includegraphics[width=0.094\textwidth]{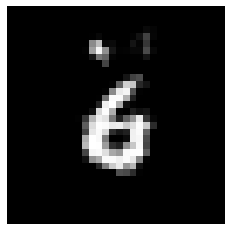}
    \includegraphics[width=0.094\textwidth]{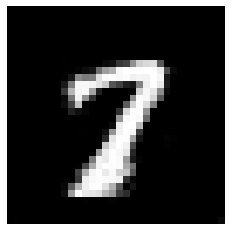}
    \includegraphics[width=0.094\textwidth]{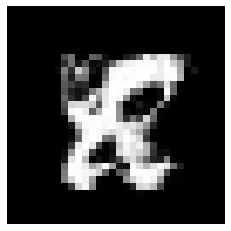}
    \includegraphics[width=0.094\textwidth]{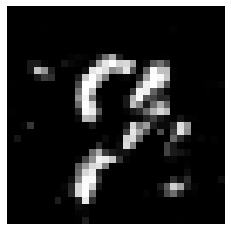}
    
    \includegraphics[width=0.094\textwidth]{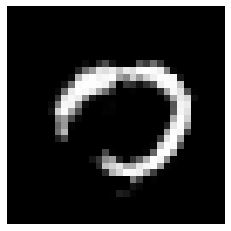}
    \includegraphics[width=0.094\textwidth]{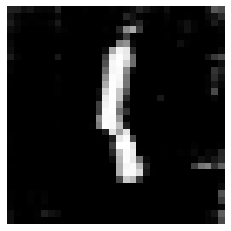}
    \includegraphics[width=0.094\textwidth]{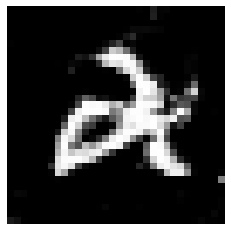}
    \includegraphics[width=0.094\textwidth]{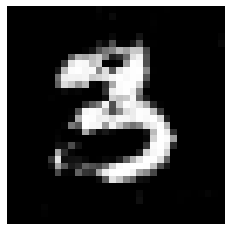}
    \includegraphics[width=0.094\textwidth]{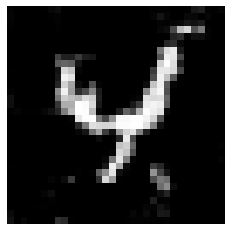}
    \includegraphics[width=0.094\textwidth]{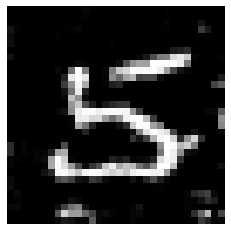}
    \includegraphics[width=0.094\textwidth]{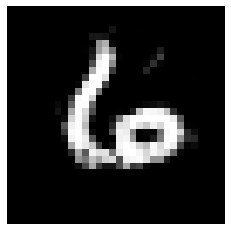}
    \includegraphics[width=0.094\textwidth]{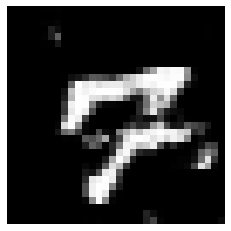}
    \includegraphics[width=0.094\textwidth]{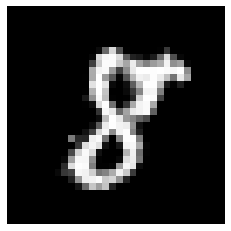}
    \includegraphics[width=0.094\textwidth]{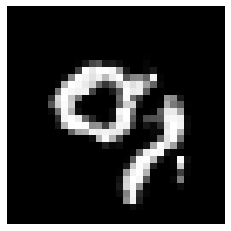}

    \includegraphics[width=0.094\textwidth]{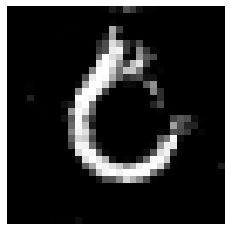}
    \includegraphics[width=0.094\textwidth]{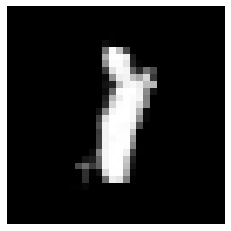}
    \includegraphics[width=0.094\textwidth]{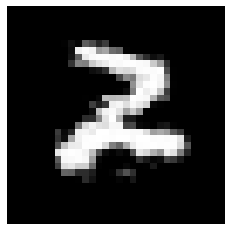}
    \includegraphics[width=0.094\textwidth]{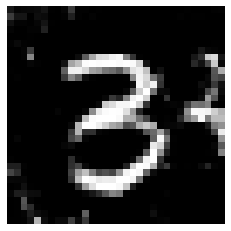}
    \includegraphics[width=0.094\textwidth]{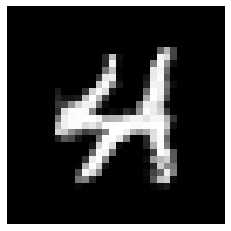}
    \includegraphics[width=0.094\textwidth]{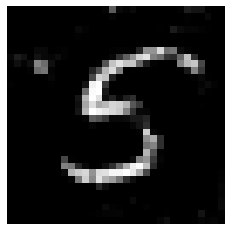}
    \includegraphics[width=0.094\textwidth]{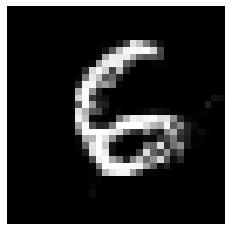}
    \includegraphics[width=0.094\textwidth]{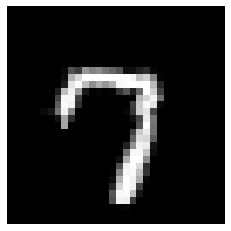}
    \includegraphics[width=0.094\textwidth]{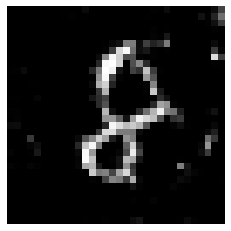}
    \includegraphics[width=0.094\textwidth]{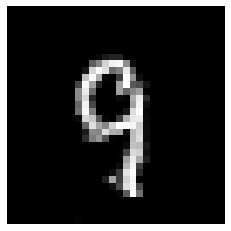}
    
    \vspace{10pt}
    
    \includegraphics[width=0.094\textwidth]{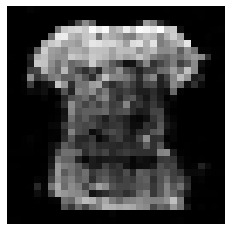}
    \includegraphics[width=0.094\textwidth]{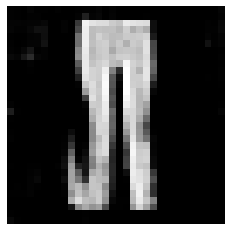}
    \includegraphics[width=0.094\textwidth]{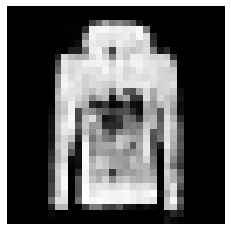}
    \includegraphics[width=0.094\textwidth]{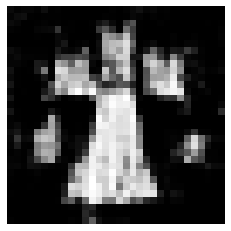}
    \includegraphics[width=0.094\textwidth]{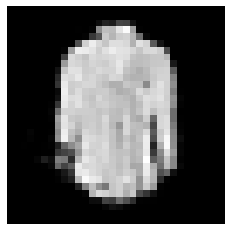}
    \includegraphics[width=0.094\textwidth]{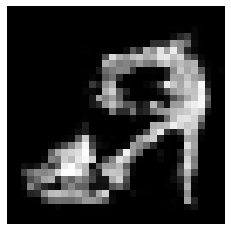}
    \includegraphics[width=0.094\textwidth]{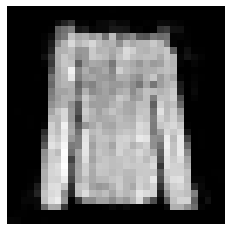}
    \includegraphics[width=0.094\textwidth]{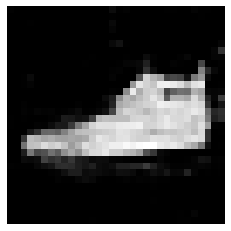}
    \includegraphics[width=0.094\textwidth]{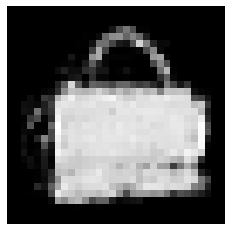}
    \includegraphics[width=0.094\textwidth]{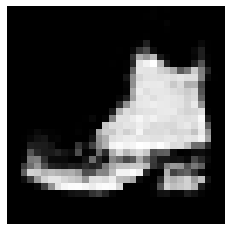}
    
    \includegraphics[width=0.094\textwidth]{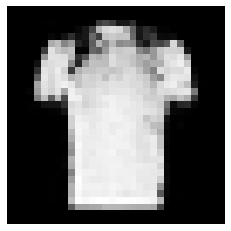}
    \includegraphics[width=0.094\textwidth]{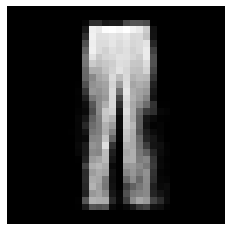}
    \includegraphics[width=0.094\textwidth]{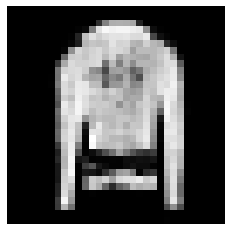}
    \includegraphics[width=0.094\textwidth]{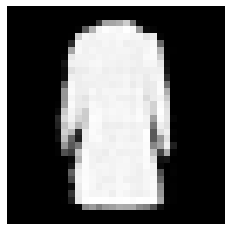}
    \includegraphics[width=0.094\textwidth]{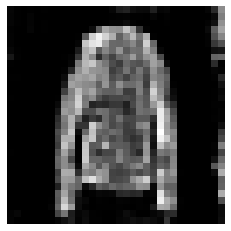}
    \includegraphics[width=0.094\textwidth]{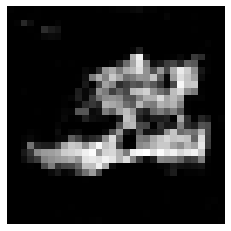}
    \includegraphics[width=0.094\textwidth]{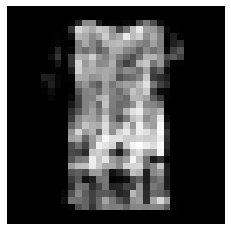}
    \includegraphics[width=0.094\textwidth]{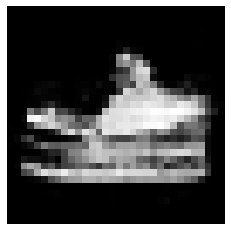}
    \includegraphics[width=0.094\textwidth]{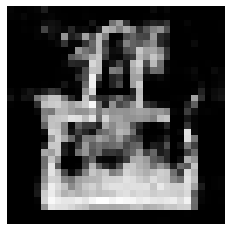}
    \includegraphics[width=0.094\textwidth]{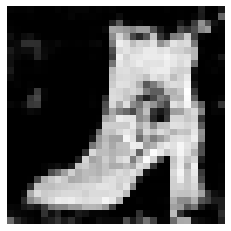}
    
    \includegraphics[width=0.094\textwidth]{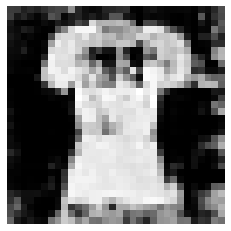}
    \includegraphics[width=0.094\textwidth]{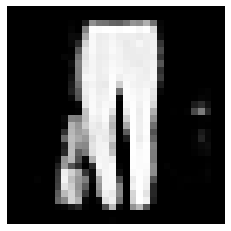}
    \includegraphics[width=0.094\textwidth]{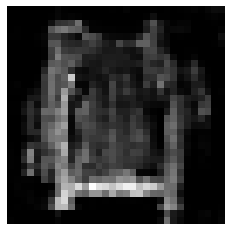}
    \includegraphics[width=0.094\textwidth]{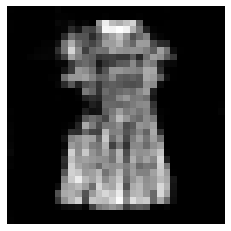}
    \includegraphics[width=0.094\textwidth]{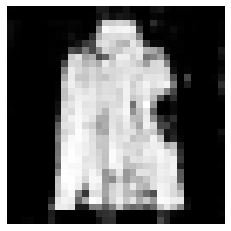}
    \includegraphics[width=0.094\textwidth]{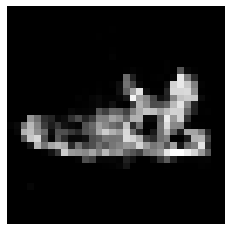}
    \includegraphics[width=0.094\textwidth]{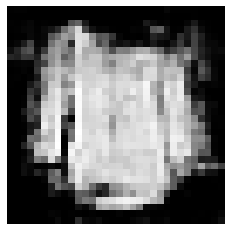}
    \includegraphics[width=0.094\textwidth]{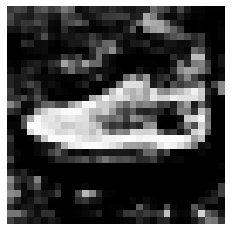}
    \includegraphics[width=0.094\textwidth]{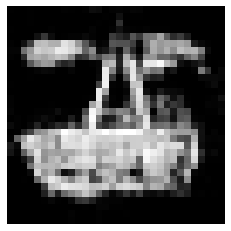}
    \includegraphics[width=0.094\textwidth]{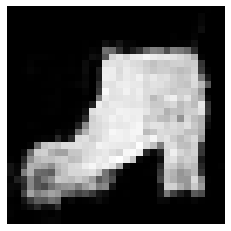}
    
    \includegraphics[width=0.094\textwidth]{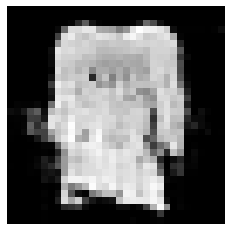}
    \includegraphics[width=0.094\textwidth]{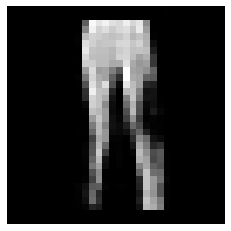}
    \includegraphics[width=0.094\textwidth]{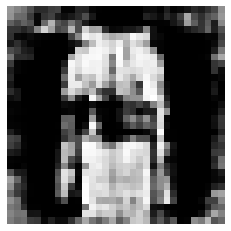}
    \includegraphics[width=0.094\textwidth]{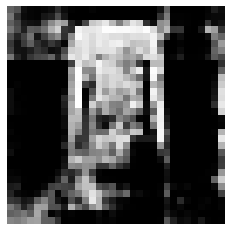}
    \includegraphics[width=0.094\textwidth]{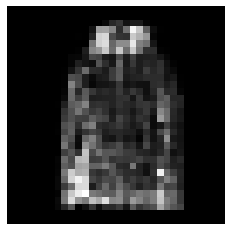}
    \includegraphics[width=0.094\textwidth]{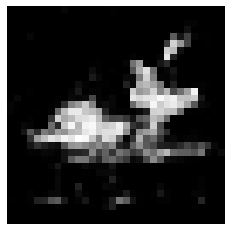}
    \includegraphics[width=0.094\textwidth]{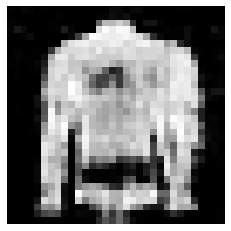}
    \includegraphics[width=0.094\textwidth]{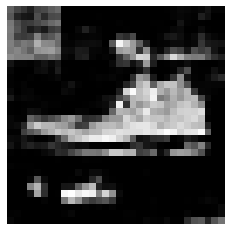}
    \includegraphics[width=0.094\textwidth]{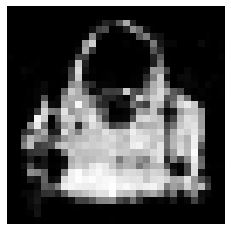}
    \includegraphics[width=0.094\textwidth]{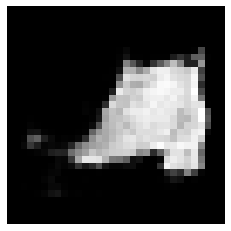}
    
  \caption{Examples of synthetic data generated from SGDE generators. The first four rows are related to MNIST, and the remaining four to Fashion MNIST. For each dataset, every column contains images from a single generator trained for that specific class. It is noticeable the presence of noise in the background and the content distortion related to the differential privacy. The images are perceptually identifiable as not real and, therefore, not related to any privacy-protected real sample from a client dataset.}
  \label{fig:synth_samples}
\end{figure*}

\subsection{Baseline Experiments}
\label{sec:baseline_exps}

The first set of baseline experiments, identified with the \emph{Local} keyword, measures the classification accuracy, F1 score, and AUC of a machine learning model over different classification tasks using 10-fold cross-validation.
Results are available in Table~\ref{tab:exps_local}.
Instead, the second set of baseline experiments measures the performance of locally-trained models on external data. 
For each client, a machine learning model is trained on the entire local dataset, evaluating its performance on the test set.
The best number of training epochs is found via cross-validation.
The results are collected in Table~\ref{tab:exps_test}.

\begin{table*}[t]
    \centering
    \caption{Experiments evaluated on local data splits. The \emph{Local} columns refer to the average performance of local models trained on local data and evaluated with 10-fold cross-validation. The \emph{Federated} columns refer to the performance of a single global model, trained with FedAvg~\cite{mcmahan_communication-efficient_2017} and evaluated on private validation splits. The \emph{Synthetic} columns refer to the average performance of local models trained with synthetic data coming from the SGDE protocol. Models in the \emph{Synthetic} columns are evaluated on the entire local datasets. We highlight the average improvement of federated learning (\emph{Federated}) and SGDE (\emph{Synthetic}) on the evaluation metrics with respect to local (\emph{Local}) training.}
    \label{tab:exps_local}
    
    \begin{tabular*}{\textwidth}{@{} l @{\extracolsep{\fill}} c @{\extracolsep{\fill}} c @{\extracolsep{\fill}} c @{\extracolsep{\fill}} c @{\extracolsep{\fill}} c @{\extracolsep{\fill}} c @{\extracolsep{\fill}} c @{\extracolsep{\fill}} c @{\extracolsep{\fill}} c @{}}
        \toprule
        \multirow{2}{*}{Dataset} & \multicolumn{3}{c}{Accuracy} & \multicolumn{3}{c}{F1 score} & \multicolumn{3}{c}{AUC}\\
        \cmidrule(l){2-4}\cmidrule(l){5-7}\cmidrule(l){8-10}
        & \emph{  Local  } & \emph{ Federated } & \emph{  Synthetic  } & \emph{  Local  } & \emph{ Federated } & \emph{  Synthetic  } & \emph{  Local  } & \emph{ Federated } & \emph{  Synthetic  }\\
        \midrule
        Titanic             & 75.67 & 76.67 & $\mathbf{80.87}$ & 19.43 & $\mathbf{69.89}$ & 63.37 & 75.70 & 69.38 & $\mathbf{78.35}$\\
        Breast Cancer       & 89.67 & 89.50 & $\mathbf{97.09}$ & 93.37 & 84.16 & $\mathbf{97.81}$ & 99.17 & 98.75 & $\mathbf{99.27}$\\
        Mushrooms           & 92.93 & 91.51 & $\mathbf{93.49}$ & 92.43 & 91.32 & $\mathbf{93.14}$ & 96.23 & 95.84 & $\mathbf{96.61}$\\
        Adult               & $\mathbf{80.64}$ & 76.01 & 79.65 & 49.69 & 47.95 & $\mathbf{61.64}$ & 83.30 & 79.81 & $\mathbf{83.73}$\\
        Wine Quality        & 93.46 & 90.44 & $\mathbf{98.54}$ & 82.98 & 85.81 & $\mathbf{97.10}$ & 99.44 & 99.10 & $\mathbf{99.49}$\\
        MNIST               & 98.20 & $\mathbf{99.40}$ & 98.72 & 98.16 & $\mathbf{99.39}$ & 98.71 & 99.02 & $\mathbf{99.66}$ & 99.31\\
        Fashion MNIST       & 88.47 & $\mathbf{91.75}$ & 89.30 & 88.32 & $\mathbf{91.65}$ & 89.22 & 93.87 & $\mathbf{95.76}$ & 94.76\\
        \midrule
        Avg. Improvement & & -0.54 & $\mathbf{+2.66}$ & & +6.54 & $\mathbf{+10.94}$ & & -1.20 & $\mathbf{+0.68}$\\
        \bottomrule
    \end{tabular*}
\end{table*}

\begin{table}[t]
    \centering
    \caption{Hyperparameters of the $\beta$-VAE architecture. For dense layers, we highlight the number of neurons, while for convolutional layers, we report the number of filters, the kernel size, and the stride value.}
    \label{tab:generators}
    
    \begin{tabular*}{\linewidth}{@{} l @{\extracolsep{\fill}} l  @{\extracolsep{\fill}} l  @{\extracolsep{\fill}} l @{}}
        \toprule
        Architecture & Layer & Tabular data & Image data\\
        \midrule
        \multirow{3}{*}{Encoder} & 1\textsuperscript{st} Layer & \texttt{Dense(64)} & \texttt{Conv2D(128,3,2)}\\
        & 2\textsuperscript{nd} Layer & \texttt{Dense(32)} & \texttt{Conv2D(256,3,2)}\\
        & 3\textsuperscript{rd} Layer & - & \texttt{Conv2D(512,3,2)}\\
        \midrule
        \multirow{3}{*}{Decoder} & 1\textsuperscript{st} Layer & \texttt{Dense(64)} & \texttt{Conv2DT(128,3,2)}\\
        & 2\textsuperscript{nd} Layer & \texttt{Dense(128)} & \texttt{Conv2DT(256,3,2)}\\
        & 3\textsuperscript{rd} Layer & - & \texttt{Conv2DT(512,3,2)}\\
        \bottomrule
    \end{tabular*}
\end{table}

\subsection{Federated Learning Experiments}
\label{sec:fl_exps}

In the set of experiments identified with the \emph{Federated} keyword, a single machine learning model is trained on a collection of private client datasets using the FedAvg algorithm~\cite{mcmahan_communication-efficient_2017}.
Table~\ref{tab:exps_local} collects the average model metrics evaluated on private validation splits from client datasets.
Instead, Table~\ref{tab:exps_test} reports the metrics of the same model evaluated on the test set. 
Federated averaging was run until convergence, imposing the average between accuracy, F1, and AUC as validation score for the early stopping criterion.

\subsection{SGDE Experiments}
\label{sec:sgde_experiments}

The experiments related to data generated using SGDE are identified with the \emph{Synthetic} keyword.
Given a specific dataset, according to the SGDE requirements, each client must build, train, and upload a data generator for each class to a trusted central server. 
Subsequently, the client can access all the available generators associated with the requested dataset.
Assuming that all the federation members are collaborative and equally interested in any available data generator, each client can access only and exclusively their private data and the set of public data generators.
At this point, clients can train machine learning models using synthetic data from all the available generators.
Thus, local machine learning models are not limited to exploit only the private data on the same device.

In the experiments marked as Synthetic, each client trains a machine learning model locally using the optimal number of generated samples produced by the generators exchanged with SGDE. The synthetic dataset is constructed from samples uniformly picked from each available generator.
In Table~\ref{tab:exps_local}, the average metrics of local models trained on synthetic samples and evaluated on private client datasets are collected under the \emph{Synthetic} column. In Table~\ref{tab:exps_test}, instead, are reported the average metrics of the same models evaluated on the test set.

\subsection{Classifiers and Infrastructure}

To ensure a fair and robust comparison, only well-known models from machine learning literature are used as classifiers.
In particular, the experiments on tabular datasets involve logistic regression as classifier, while the ones on image datasets include the first eight pre-trained layers of VGG16~\cite{simonyan2014very} adapted with transfer learning. In the latter case, the only trainable component of the VGG16 architecture is the last 256-neuron dense layer with LeakyReLU as activation function, followed by the Softmax classifier. VGG16 is trained using the Adam optimizer with a learning rate of 0.001. All experiments are executed on a system equipped with an Intel(R) Xeon(R) CPU E5-2630 v4 @ 2.20GHz and an Nvidia Quadro RTX 6000 GPU.

\subsection{Differentially Private Generators}
\label{sec:betavaes}

\begin{table*}[t]
    \centering
    \caption{Experiments evaluated on test sets. The \emph{Local} columns refer to the average performance of local models trained on local data. The \emph{Federated} columns refer to the performance of a single global model, trained with FedAvg~\cite{mcmahan_communication-efficient_2017}. The \emph{Synthetic} columns refer to the average performance of local models trained with synthetic data coming from the SGDE protocol. All the models are evaluated on a hold-out set. We highlight the average improvement of federated learning (\emph{Federated}) and SGDE (\emph{Synthetic}) on the evaluation metrics with respect to local (\emph{Local}) training.}
    \label{tab:exps_test}
    
    \begin{tabular*}{\textwidth}{@{} l @{\extracolsep{\fill}} c @{\extracolsep{\fill}} c @{\extracolsep{\fill}} c @{\extracolsep{\fill}} c @{\extracolsep{\fill}} c @{\extracolsep{\fill}} c @{\extracolsep{\fill}} c @{\extracolsep{\fill}} c @{\extracolsep{\fill}} c @{}}
        \toprule
        \multirow{2}{*}{Dataset} & \multicolumn{3}{c}{Accuracy} & \multicolumn{3}{c}{F1 score} & \multicolumn{3}{c}{AUC}\\
        \cmidrule(l){2-4}\cmidrule(l){5-7}\cmidrule(l){8-10}
        & \emph{  Local  } & \emph{ Federated } & \emph{  Synthetic  } & \emph{  Local  } & \emph{ Federated } & \emph{  Synthetic  } & \emph{  Local  } & \emph{ Federated } & \emph{  Synthetic  }\\
        \midrule
        Titanic         & 71.83 & $\mathbf{74.81}$ & 74.01 & 29.70 & $\mathbf{71.61}$ & 56.00 & 77.14 & 70.85 & $\mathbf{77.43}$\\
        Breast Cancer   & 89.42 & 91.86 & $\mathbf{93.02}$ & 92.25 & 90.82 & $\mathbf{94.78}$ & 99.60 & 99.36 & $\mathbf{99.76}$\\
        Mushrooms       & 92.56 & 91.27 & $\mathbf{93.49}$ & 91.92 & 91.20 & $\mathbf{93.14}$ & 96.30 & 96.07 & $\mathbf{96.61}$\\
        Adult           & $\mathbf{80.87}$ & 76.47 & 79.00 & 50.14 & 47.70 & $\mathbf{60.21}$ & 84.02 & 81.24 & $\mathbf{84.08}$\\
        Wine Quality    & 92.57 & 90.23 & $\mathbf{97.79}$ & 82.42 & 85.10 & $\mathbf{95.70}$ & 98.63 & 98.50 & $\mathbf{98.65}$\\
        MNIST           & 97.76 & $\mathbf{99.08}$ & 98.49 & 97.71 & $\mathbf{99.08}$ & 98.49 & 99.02 & $\mathbf{99.50}$ & 99.19\\
        Fashion MNIST   & 85.97 & 87.94 & $\mathbf{88.13}$ & 85.81 & 87.99 & $\mathbf{88.04}$ & 92.65 & 93.68 & $\mathbf{94.13}$\\
        \midrule
        Avg. Improvement & & +0.10 & $\mathbf{+1.85}$ & & +6.22 & $\mathbf{+8.06}$ & & -1.17 & $\mathbf{+0.36}$\\
        \bottomrule
    \end{tabular*}
\end{table*}

Data generators are the core component behind SGDE, as they need, on the one hand, to guarantee strict security levels and, on the other hand, to produce valuable synthetic data.
In this sense, generated samples do not need to produce perceptively realistic data as long as they provide high utility in machine learning tasks. 
In fact, as shown in Figure~\ref{fig:synth_samples}, which depicts some synthetic images coming from class-specific generators of a single client, subjects are quite noisy and do not closely resemble any real sample. Nevertheless, the experimental results that will be presented in Section~\ref{sec:results} highlight a high machine learning utility for synthetic images, sometimes even higher than real data.

Concerning the level of security, data generators must not leak information related to the private data used during their training phase, thus preserving client privacy.
SGDE allows the sharing of any data generator that meets the security requirements. 

In our experiments, we implemented a custom version of the $\beta$-VAE~\cite{higgins2016beta} architecture trained with the differentially private implementation of the Adam optimizer~\cite{abadi_deep_2016}.
In order to achieve the highest resilience level from inference attacks, i.e., the attack is not more effective than random guessing~\cite{rahman2018membership,hayes2019logan}, each client must train its generators so that their final $(\varepsilon,\delta)$-DP level is characterized by $\varepsilon \leq 1.5$, $\delta \ll \frac{1}{|\mathcal{D}_c|}$ and $RDP \geq 9$, where $|\mathcal{D}_c|$ is the number of samples belonging to class $c$ of dataset $\mathcal{D}$ and $RDP$ is the R\'{e}nyi-DP value. Each member must cut off from every model its initial part, the encoder, and share only the portion meaningful to generate synthetic samples, the decoder.

The ablation study conducted on the hyperparameters of the $\beta$-VAE architecture maximizes the generation performance against the noise introduced with differential privacy.
The ablation study consists of a hyperparameter grid search on two architectural configurations for tabular and image data.
The results are shown in Table~\ref{tab:generators}. 
In the model architecture for tabular data, each dense layer is followed by a LeakyReLU activation function. 
Instead, in the model architecture for the image data, each convolutional layer is followed by a Swish~\cite{ramachandran2017searching} activation function.
Finally, the latent space dimension and the $\beta$ value are tuned for each dataset.

\subsection{Results}
\label{sec:results}

From an individual client perspective, the most compelling question is whether joining the SGDE protocol is beneficial from a utility standpoint.
In Table~\ref{tab:exps_local}, each member of the federated network improved the classification average accuracy and AUC by 2.66\% and 0.68\%, respectively over all the datasets. 
This result allows us to assert that synthetic data coming from generators exchanged with SGDE is more effective than single-client local data to learn a classification task.
Thus, synthetic data can effectively substitute privacy-protected local data in a machine learning procedure.
In fact, in our experiments, SGDE is crucial to increase the amount of available information of a single client to train a machine learning model, without compromising the privacy of the other federation individuals.

Moreover, the F1 score improves by 10.94\% on average over all the experiments.
This means that generating data from each class uniformly lowers the distribution bias with respect to underrepresented classes in unbalanced datasets, as each client has access to more information about minority classes.
Therefore, taking part to the SGDE protocol and training a machine learning model on synthetic data is not only beneficial from the accuracy standpoint, but leads to a more fair classification performance overall.

The results are confirmed in Table~\ref{tab:exps_test} too, where models are evaluated on the test set. 
By combining the \emph{Local} and \emph{Synthetic} experiments, there is an average improvement of accuracy and AUC of 1.85\% and 0.36\%, respectively.
The F1 score increases by 8.06\% on average, confirming the strong fairness advantage granted by taking part in the SGDE generator exchange.

So far, the discussion focused on the performance difference of training on synthetic samples with respect to training on real local data.
Then, the next natural question is how training on synthetic samples compares to federated learning, the most prominent privacy-preserving training technique for a machine learning model on distributed data.
The interesting result is that in most cases SGDE has still an advantage, not only by design in transparency communication costs, but in the final classification performance too. 
The experiments show that SGDE performs similarly, or even outperforms the standard FedAvg algorithm~\cite{mcmahan_communication-efficient_2017} in settings with a small number of clients and unbalanced data distributions.
The result is more evident in the first five rows of Table~\ref{tab:exps_local} and Table~\ref{tab:exps_test}, where the experiments involve small tabular datasets and logistic regression as global machine learning model. 

To recap, the experiments showed that SGDE provides a secure way of sharing knowledge embedded in private data by collecting data generators in a single pool and making them publicly accessible.
With these results, SGDE has proven capable of improving the performance of machine learning tasks of individuals taking part in the generators exchange.
Moreover, as SGDE allows the generation of a transparent, local dataset, it eliminates the need to join an iterative model exchange procedure, as in federated learning, alleviating the communication overload.
Additionally, SGDE is beneficial from the accuracy, fairness, and transparency standpoints with respect to standard FL, while still guaranteeing strong protection for private user data.

More generally, we advocate for the development of secure technologies based on publicly accessible synthetic data, as we believe individuals cooperation in a secure environment to be the key to increasing value and knowledge in a privacy-compliant manner.

\section{Conclusion}
\label{sec:conclusion}

This work presents SGDE, a secure data exchange framework based on data generators with high privacy guarantees. 
The benefits of a generative-centric approach to data sharing are extensively discussed in a context where granting privacy is a hard constraint.
Generators retain the distinctive features of private data while providing access to an arbitrarily large set of synthetic samples that are public, reproducible, and fair.
Moreover, a centralized dataset is resilient against poisoning and inference attacks which pose a real threat to standard federated learning.

The effectiveness of SGDE is showcased in several experimental scenarios with high confidentiality levels, employing differentially private $\beta$-VAEs as generators. 
Training on synthetic data yielded better performance than true, private data, granting privacy protection for the individuals.
SGDE consistently outperformed federated learning, one of the most influential techniques to train a machine learning model in a privacy-preserving way from distributed data. In fact, the inclusion of a generative protocol to share privacy-compliant information is more communication-efficient, transparent, and effective than iteratively training a model with gradients exchange, especially when data distributions are skewed among clients.

Today, many researchers praise the benefits of a generative approach in privacy-critical federations. We believe that this is a research direction worth exploring, with a strong potential to lower the obstacles towards more user-centric, fair, and secure large-scale machine learning.

\section*{Acknowledgement}
The European Commission has partially funded this work under the H2020 grant N. 101016577 AI-SPRINT: AI in Secure Privacy-pReserving computINg conTinuum.

\bibliographystyle{IEEEtranS}
\bibliography{biblio}
\end{document}